# BIM Hyperreality

**Mohammad Alawadhi**
Texas A&M University

**Wei Yan**
Texas A&M University

## Data Synthesis Using BIM and Hyperrealistic Rendering for Deep Learning

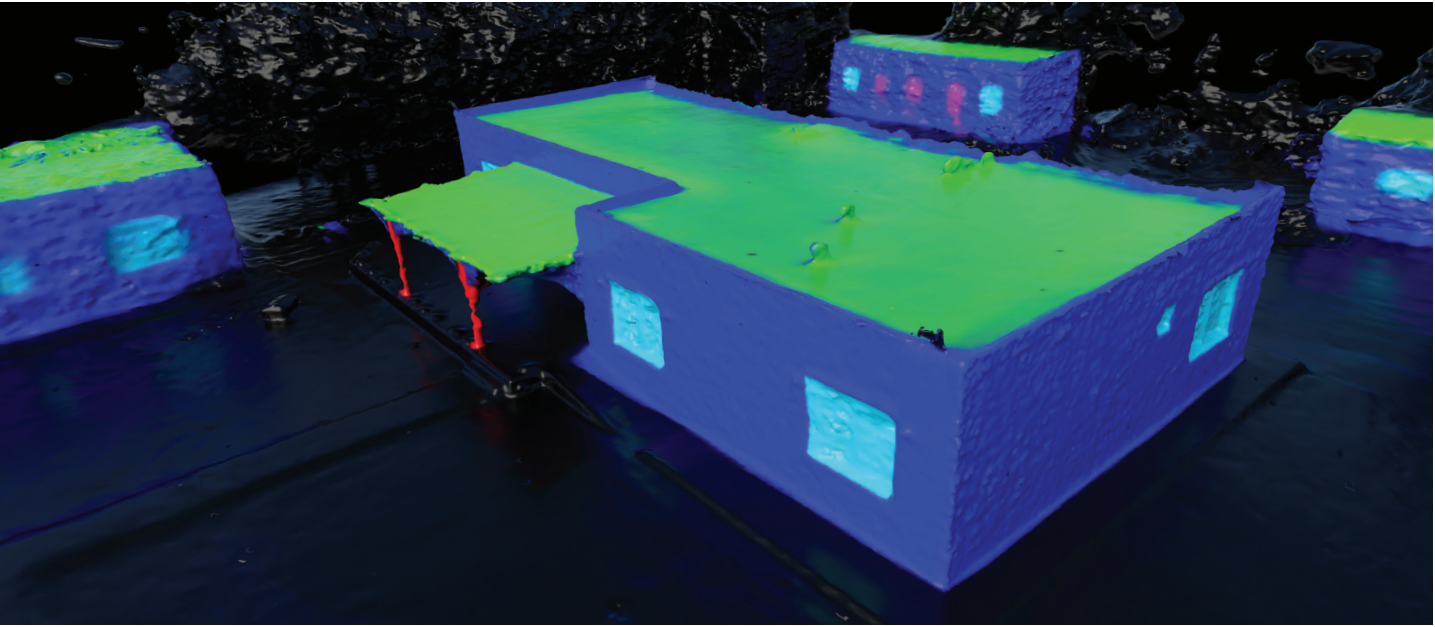

### ABSTRACT

Deep learning is expected to offer new opportunities and a new paradigm for the field of architecture. One such opportunity is teaching neural networks to visually understand architectural elements from the built environment. However, the availability of large training datasets is one of the biggest limitations of neural networks. Also, the vast majority of training data for visual recognition tasks is annotated by humans. In order to resolve this bottleneck, we present a concept of a hybrid system—using both building information modeling (BIM) and hyperrealistic (photorealistic) rendering—to synthesize datasets for training a neural network for building object recognition in photos. For generating our training dataset—*BIMrAI*, we used an existing BIM model and a corresponding photo-re-alistically rendered model of the same building. We created methods for using renderings to train a deep learning model, trained a generative adversarial network (GAN) model using these methods, and tested the output model on real-world photos. For the specific case study presented in this paper, our results show that a neural network trained with synthetic data; i.e., photorealistic renderings and BIM-based semantic labels, can be used to identify building objects from photos without using photos in the training data. Future work can enhance the presented methods using available BIM models and renderings for more generalized mapping and description of photographed built environments.

1　3D model generated using photos and a render-trained GAN.

## INTRODUCTION

The inability of a cognitive intelligence—either human or artificial—to distinguish between reality and simulated reality, has been coined as *hyperreality* in philosophical discourse, first by Baudrillard (1995). Humans as much as computer machines can struggle to differentiate between a photo and an imitation of a photo—either as art or a 3D rendering. This argument is evident in hyperrealistic building visualizations (digitized architecture) and counterpart actualized buildings (built architecture) that are common practice in the architecture, engineering, and construction (AEC) industry, which are being augmented by a growing interest in virtual reality and augmented reality. By considering hyperrealism with the proliferation and advancement of artificial intelligence technology for solving machine cognition tasks, the authors' work was an attempt to exploit hyperreality between synthetic photorealistic renderings and photos of buildings for teaching machines to detect building objects—as defined through BIM. This attempt constituted: (1) exploring methods to create machine learning datasets that rely on BIM and photorealistic rendering instead of manual annotation, and (2) testing said methods using current neural network technology.

The introduction of artificial neural networks into architectural discourse is a recent phenomenon owing itself to the recent breakthroughs of *deep* neural networks at solving complex machine learning tasks, paving the way to a new paradigm of *deep learning*. Neural networks are inspired by biological neurons, in a sense, they are an abstract model of the brain. Their typical structure is composed of an input layer of nodes, hidden layers of nodes, and an output layer with a single node, with connections between them that have weights, which are multiplied and adjusted. They require training by showing them many examples. An area where deep neural networks has excelled is in image processing tasks, which include image classification, object detection, semantic segmentation, image-to-image translation, and image generation, among other tasks. Since architecture is a visually-oriented field, a significant portion of previous work was interested in using these image-based deep learning models within the field. However, few have explored using the knowledge contained in digital BIM models to train deep learning models.

In their work, Krijnen and Tamke (2015) followed by Kim et al. (2019) used BIM-based data to train neural networks for BIM model quality assurance and data integrity. As et al. (2018) developed a methodology to train neural networks using BIM-based graph models and subjective scores. The previously mentioned literature did not explore the applications of BIM-trained neural networks in the real built environment; an area this paper is interested in. Braun and Borrmann (2019) proposed using camera-aligned BIM models and photos of construction sites to train neural networks, though no training was conducted. This paper is particularly interested in the use of synthetic data to train a neural network that can be used on real-world data (i.e., photos), when given an example photo of a building, a trained network would segment the image into predicted building objects. This application would be useful for building surveys and building energy modeling (BEM).

The potential of neural networks is limited by the availability of large, high-quality training data. Also, web data and manual annotation are the dominant data acquisition methods in visual recognition (Mayer et al. 2018). A solution idea for this data acquisition problem is using both BIM and photorealistic rendering for generating training data. By considering the ubiquity of BIM and 3D rendering in the architecture field, BIM can generate building object data that can be used with photorealistic renderings to train a neural network.

3D renderings can be impossible to differentiate from photos, and even computer programs can struggle to differentiate between them (Lyu and Farid 2018). Synthetic data—in the form of 3D renderings—were used for training machine learning models to understand indoor architectural scenes (Georgakis et al. 2017; Handa et al. 2015; Peng et al. 2017; Zhang et al. 2017). Machine learning models trained using a dataset supplemented by photorealistic renderings can outperform programs trained with only hand-labeled photos (Richter et al. 2016). Photorealistic renderings also reduced the number of manually-annotated photos required for training (Georgakis et al. 2017; Richter et al. 2016). Neural networks trained on photorealistic renderings can accurately estimate depth and part segmentation in photos (Varol et al. 2017) and can produce results near or beyond state-of-the-art performance (Handa et al. 2015; Zhang et al. 2017).

While previous research in other fields has explored the use of synthetic data in the form of annotated photorealistic renderings for training neural networks that can predict objects from photos, there is no example that utilized this in conjunction with BIM—an object-oriented process. Therefore, it can be hypothesized that a neural network trained with BIM and photorealistic renderings of buildings would work well when tested on photos of buildings. Proving this hypothesis true would mean that available BIM models and visualization renderings that are already being used in the AEC industry can be leveraged instead of manually labeling building photos.

For testing the previously discussed hypothesis, Conditional GANs (CGANs)—a variation of GANs—can be trained with RGB image pairs of BIM renderings and photorealistic renderings for semantic segmentation of building objects. GANs are an emerging type of neural network introduced by Goodfellow et al. (2014) which is based on implementing two neural networks competing in a zero-sum game (Goodfellow 2017). These neural networks are set up where one network is a generator that creates samples with the same distribution as the training data, while another network is a discriminator that determines whether the samples are real or fake (Goodfellow). The generator network is trained to deceive the discriminator (Goodfellow). GANs have a large potential in the future of machine learning technology. CGANs are successful solutions for image processing problems that involve translating an input image into a corresponding output image (Isola et al. 2018); for example, a building façade image and a corresponding labeled façade image. GANs, as opposed to state-of-the-art semantic segmentation neural networks, can be used for semantic segmentation problems in addition to other image-to-image translation tasks, which demonstrate their applicability and flexibility for architectural problems, including generative design (Chaillou 2019), urban image generation (Steinfeld 2019), depth estimation, and detecting architectural elements from photos.

Utilizing GANs for achieving machine understanding of basic objects of the built environment can be useful for a multitude of tasks in real-world conditions. For example, a neural network that is trained to predict basic building objects from photos can be used on building photos that are acquired from photogrammetry surveys, as one of the challenges of automated BEM from photogrammetry is the annotation of basic building-element objects such as windows. Automating the annotation process using machine learning can be used instead of manual annotations; therefore, a photogrammetry-based building model annotated using a neural network can be useful for producing object-oriented building models including BEM and BIM models.

In the scope of our work, we tackle the issue of manual annotation in deep learning by developing methods for data synthesis derived from BIM data and photorealistic rendering. We experimented with a single-sample test case of a building that includes on-site photos, a corresponding BIM model, and a photorealistic 3D model of that building. Then, we used *BIMrAI*—the rendering output of our data synthesis system—to train a neural network. We evaluated the testing results on photos in terms of how well the network performed and analyzed the methods in terms of potential usefulness for mapping the built environment and automatic BIM creation. The following section describes the methods that were developed.

## METHODS

A system was developed by the authors using BIM and 3D software tools to generate synthetic training data for a neural network model that conducts semantic segmentation on photos. The network model that was chosen for testing the efficacy of the system is *pix2pix*—a state-of-the-art CGAN developed by Isola et al. (2018) for image-to-image translation: meaning, it translates an input RGB image to a corresponding output RGB image. In this case, the network was trained to translate photorealistic renderings to color-ID renderings—i.e., the network was adapted for the task of semantic segmentation, which is segmenting an input image into color-labeled objects. This approach was used to conduct semantic segmentation based on building objects, where each color label corresponds to a building object (Figure 4).

The system was tested with a real building sample: a single-story farmhouse with an L-shaped floor plan including 5 surrounding buildings. The synthetic training data was generated using two 3D models: a BIM model, and a corresponding photorealistic model—based on photo-grammetry—of the same building that was acquired from previous work (Alawadhi and Yan 2018). It is possible to generate the data using a single BIM model with photorealistic materials; however, this approach was not taken in the scope of this experiment.

Setting up the system for data synthesis is a four-step process: (1) assigning BIM object colors, (2) running scripts for model alignment and (3) the orbiting camera, and (4) rendering. The workflow process of this system is as follows: Using 3ds Max—a 3D software tool, an imported Revit model of the sample was aligned to the photorealistic model. The alignment was conducted using a three-point-click method. The imported BIM objects; namely, the walls, windows, doors, columns, and roofs were assigned specific user-defined RGB colors, which are blue, cyan, purple, red, and green, respectively. An animated camera that orbits around the models was automatically created using param-eters (3D positions and focal length) derived from drone camera data in the photogrammetry file.

Figure 2 shows two instances of the user interface (UI) illustrating camera positions and matching views to the aligned models where (a) is the photorealistic model and (b) is the imported BIM model (with the photorealistic model hidden).

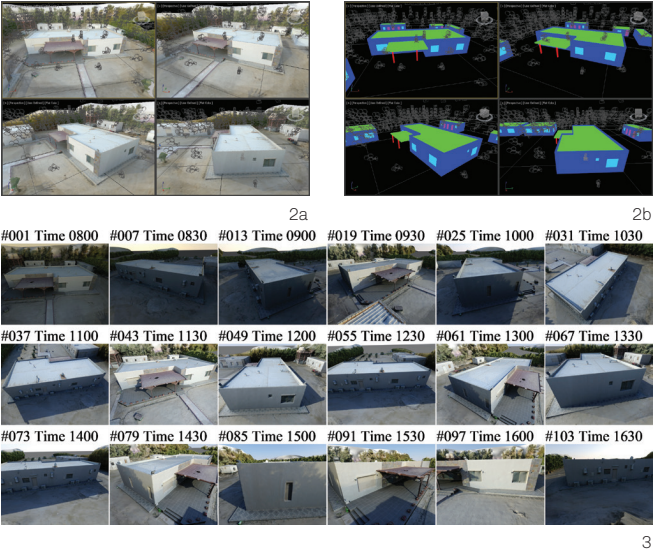

2a

2b

#001 Time 0800   #007 Time 0830   #013 Time 0900   #019 Time 0930   #025 Time 1000   #031 Time 1030

#037 Time 1100   #043 Time 1130   #049 Time 1200   #055 Time 1230   #061 Time 1300   #067 Time 1330

#073 Time 1400   #079 Time 1430   #085 Time 1500   #091 Time 1530   #097 Time 1600   #103 Time 1630

2   UI instances showing the photo-realistic model (a) and the BIM model (b).

3   Batch-rendered views with simulated lighting for each scene state.

The camera has a keyframe on each second of the animation timeline, where each keyframe represents a different camera view of the models for a total of 110 views. 38 different scene states were set up for the photorealistic model to simulate lighting conditions based on location and time of day in order to simulate realism in the training data. Figure 3 shows render samples of the photorealistic model under different conditions, where each image represents a time of day with different illumination and shadows. Given the large variations that can be produced with different views and lighting, the system inherently applies data augmentation that is often applied to training data. Another scene state was set up for the imported and color-labeled BIM model. As outputs, photorealistic renderings and corresponding color-labeled renderings were produced automatically via a batch rendering process for all 39 scene states using 3ds Max. In total, 4,180 image pairs of matching photorealistic views and color-labeled model views were rendered in 3 days using a single PC. For future work, this duration can be drastically reduced using a render farm.

The output renderings of this system were processed for use as training data for the neural network. Each rendered pair was resized to two squares and stitched together—as 256-pixel square input and output—using Python scripting for training the GAN for translating the photorealistic renderings into the BIM-based object color renderings.

Figure 4 shows the basic structure of the GAN. Given a latent vector $z$, input photorealistic rendering $c$, and output color-labeled rendering $x$, the generator network $G(c,z)$ is trained to generate the RGB color labels from the photorealistic renderings $c$ to fool the discriminator network $D(c,x)$. While the discriminator network $D(c,x)$ is trained to detect the correct labels $x$ from the false labels. There is also the L1 loss which is a regression problem between the generator's labels and the original labels $x$. The L1 loss can be disregarded in other GAN applications but is considered useful for image segmentation tasks (Isola et al. 2018); therefore, it was included in this experiment. The full loss equation in Figure 4 shows the relationship between the three loss functions: the generator loss, the discriminator loss, and the L1 loss.

Table 1 shows the main hyperparameters for training the GAN model. Total training time on the BIMrAI dataset was approximately 15 hours using a single GPU. After training, the network was tested on real-world photos of the sample building. Unlike conventional GANs in which photo images are used as training data in the GAN, in this work, only the photorealistic renderings and BIM-based color labels were used as training data. The purpose is twofold: (1) to explore if generated synthetic imagery can be used as training data for deep learning, and (2) whether the results are adequate or not.

**Table 1: GAN Training Parameters**

| Dataset Size (render pairs) | Model Resolution (pixels) | Hyperparameters (pix2pix) | | |
|---|---|---|---|---|
| | | Batch Size | Learning Rate | Training Epochs |
| 4,180 | 256 x 256 | 1 | 0.0002 | 200 |

As for evaluating the results of the trained GAN on the photos, besides evaluating the results subjectively, it is difficult to evaluate GANs objectively (Borji 2018; Salimans et al. 2016) as opposed to conventional segmentation networks. However, since the GAN was used for the same task, an evaluation method was developed to plot the accuracy based on standard metrics that are used to evaluate conventional segmentation networks, such as per-object pixel accuracy and intersection-over-union (IoU). The first step of this evaluation method is to apply a color quantization algorithm on the predicted outputs of the network for reducing the RGB colors to match the original RGB integer values that were defined in the training data synthesis system. Afterward, an evaluation tool was developed using Python. The ground truth renderings and the color-quantized GAN outputs were encoded as numerical labels corresponding to pixels and their object class; e.g., wall, window, etc. The evaluation method might be detrimental to the evaluation scores in cases where GAN-predicted RGB pixels do not correspond to the user-defined object class's RGB integers, possibly resulting in mislabeled pixels after

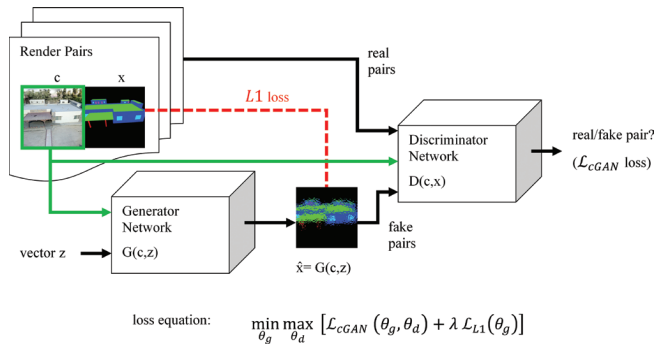

loss equation:   $\min_{\theta_g} \max_{\theta_d} \left[ \mathcal{L}_{cGAN}(\theta_g, \theta_d) + \lambda \, \mathcal{L}_{L1}(\theta_g) \right]$

4    CGAN model. Source: Figure adapted from Herranz (2018).

color quantization and label encoding. However, this is not a concern if the network has been trained well, as the pixels with observable RGB color inconsistencies introduced by the GAN would be nonexistent or very small. Also, this did not appear to be a problem in this experiment. Therefore, the presented method can be used for evaluating semantic segmentation results of the GAN objectively while identifying that a technical bias might arise in future work.

The evaluation tool uses the binary classification of object classes per-pixel. The classes are *wall*, *window*, *door*, *column*, *roof*, and *background* objects. The evaluation metrics in the tool are the accuracy, precision, recall, F1 score, and the mean IoU. The accuracy is calculated by dividing the sum of true positives *tp* and true negatives *tn*—which are predicted pixels that have the same positive and negative classifications as the ground truth—by the total number of pixel classifications. Accuracy was calculated globally (for all pixels) and for pixels per-object. The precision, recall, and F1 score take the false positives *fp* and false negatives *fn* into consideration for more in-depth analysis and are given by the following:

$$precision = \frac{tp}{tp + fp}, \quad recall = \frac{tp}{tp + fn}, \quad F1 = 2 \cdot \frac{precision \times recall}{precision + recall}$$

The mean IoU is another important metric for gauging the overlap between ground truth *a* and predicted *b* segmentations. IoU is scored from 0 to 1, where 1 indicates matching prediction and ground truth segments. The IoU is calculated for each class then averaged between all classes and is given by the following:

$$IoU(a, b) = \frac{a \cap b}{a \cup b} = \frac{tp}{tp + fp + fn}$$

The intersection between *a* and *b* is synonymous with *tp*, and the union between *a* and *b* is the sum of *a* and *b* segmentations. The mean IoU is considered a more rigorous metric than accuracy for segmentation tasks because it is sensitive to class imbalance in the data, whereas using accuracy only could give misleadingly high scores if class imbalance is an issue. Therefore, both accuracy and mean IoU were considered when evaluating the GAN results.

## RESULTS AND REFLECTION

The trained GAN was tested with real-world photos of the sample. Based on qualitative observation and comparing GAN predictions with the ground truth, the trained network was able to predict objects from the photos with acceptable accuracy—with minor misidentifications (Figure 5). Overfitting can be an issue if the training data (renderings in the test) is too similar to the testing data (photos). In the case of this test, there are noticeable differences between the renderings and the photos when compared side-by-side. Specifically, there are differences between simulated and real geometry, lighting, colors, and details when compared by human eyes. However, more future work will be needed to test if this network would work well on photos of another building sample. Most likely, the network would need to be trained on more buildings.

Therefore, what can be concluded from this test is that rendering can be generalized as a data synthesis method for deep learning, although more future work will be needed for data synthesis using BIM and rendering methods to achieve improved generalization, which is the ability to perform object recognition on other buildings for which we do not possess BIM models using the trained network. Neural networks that are trained using a large amount of data with a good amount of variance within a specific data distribution have shown to have good generalization when deployed on unseen data. For example, a neural network for predicting building objects that is trained with a large number of building images will most likely perform accurately on other unseen building images that are similar in typology. However, there is a bottleneck in data acquisition, as curating training data for deep learning is a labor-intensive task. The results show promise in tackling this issue through data synthesis using BIM and photorealistic rendering, both of which are ubiquitous in the field of architecture.

As for analyzing the results quantitatively, the GAN outputs were evaluated empirically using the evaluation tool, and the resulting metrics are shown in Table 2. The average accuracy is 93% (rounded) and the mean IoU is 0.64, which are good compared to the results from state-of-the-art benchmarks on street scene datasets used for autonomous driving applications (Table 3). For this specific building case study, the results demonstrate that a GAN can be trained using only photorealistic renderings and BIM-based labels to identify building elements from photos.

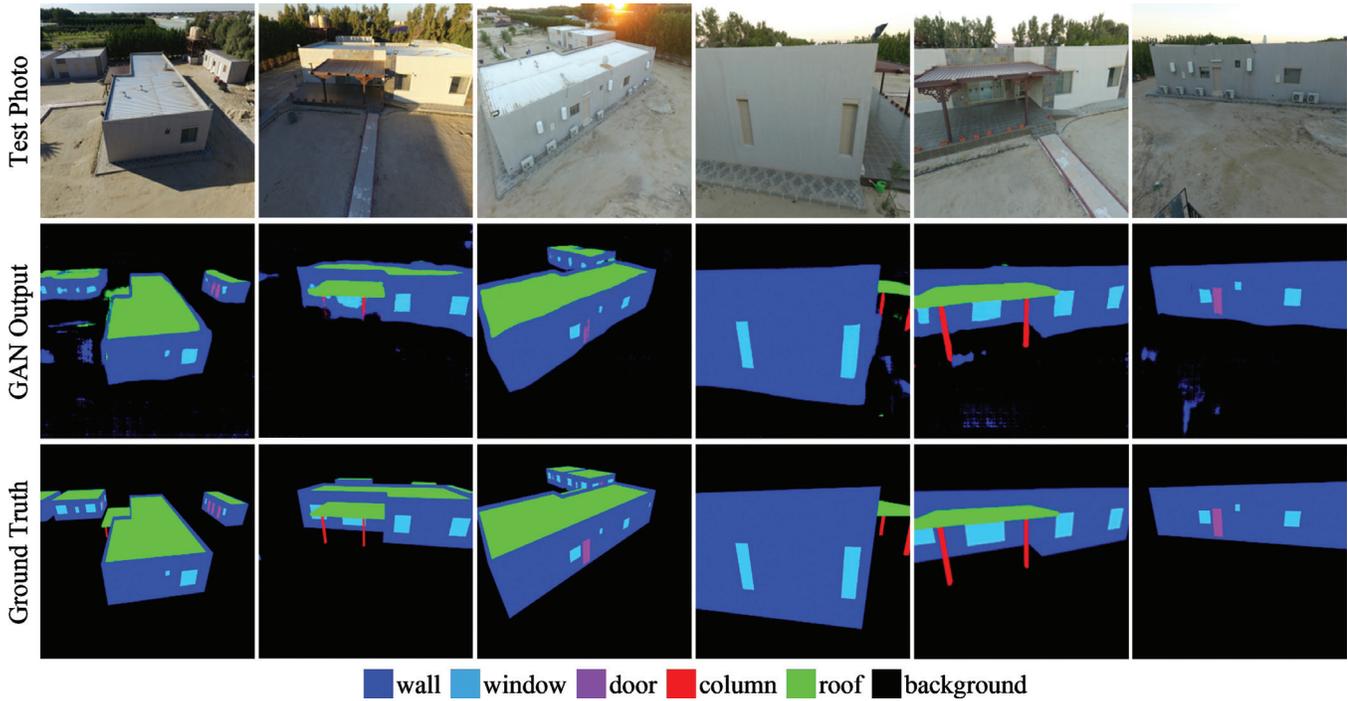

| | | | | | |

**wall** **window** **door** **column** **roof** **background**

5   Predicted GAN outputs from photo inputs.

**Table 2: GAN Semantic-Segmentation Evaluation**

| | |
|---|---|
| Accuracy (%) | **92.82** |
| Per-Object Accuracies (%) | |
| Wall | 93.69 |
| Window | 85.81 |
| Door | 55.83 |
| Column | 88.89 |
| Roof | 76.36 |
| Background | 94.49 |
| Precision | 0.934 |
| F1 | 0.926 |
| mIoU | **0.641** |

**Table 3: Results Comparison with Street Scene Benchmarks**

| Dataset | Model | Accuracy (%) | mIoU | Source |
|---|---|---|---|---|
| CamVid | FC-DenseNet103 | 91.5 | 0.669 | Jégou et al. 2017 |
| | BiSeNet | - | 0.687 | Yu et al. 2018 |
| Cityscapes | PSPNet | - | 0.802 | Zhao et al. 2017 |
| | BiSeNet | - | 0.789 | Yu et al. 2018 |
| | pix2pix | 86 | 0.35 | Isola et al. 2018 |
| | CycleGAN | 58 | 0.16 | Zhu et al. 2020 |
| GTAV-to-Cityscapes | CBST | - | 0.47 | Zou et al. 2018 |
| | ADVENT | - | 0.455 | Vu et al. 2019 |
| **BIMrAI (Ours)** | pix2pix | **92.82** | **0.641** | This Paper |

Further, to study and speculate on potential applications of the trained network, the predicted outputs from the GAN were automatically remapped onto a photogrammetry-based model. A photogrammetry software was used for the remapping process. The software normally uses photos for both the geometry and texture creation; however, for the texture, the photos were replaced with the GAN predictions obtained from the previous phase. The resulting *semantic mesh model*—i.e., the model with predicted outputs of the network projected in 3D—can be seen in Figures 1 and 6. This semantic mesh model can be used in the future as a basis to construct a BIM model from a photogrammetry model, which is in great need for applications such as BEM, building operation, renovations, etc.

An observation from this experiment is that remapping the predicted images on the 3D model seemed to increase the fidelity of the predicted colors, as incorrect predictions were blended away with correct predictions from different image angles. The blended predictions benefitted from the high average accuracy of the GAN to rectify single-image prediction colors in photo angles where the network underperformed. Figure 7 shows the increased prediction accuracy from the photogrammetry texture creation process when compared to single-image prediction. Photogrammetry tools implement texture mapping algorithms using Bayesian inference methods, such as Markov random field for high-quality texture blending (Lempitsky and Ivanov 2007). The blended prediction method (Figure 6)

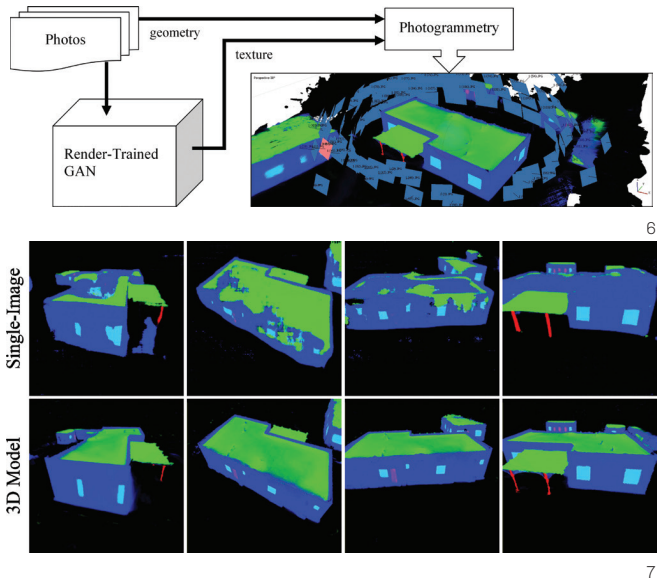

Single-Image

3D Model

6    Process for generating semantic mesh model toward BIM creation.

7    Single-image vs. 3D-projected and blended predictions.

has the potential to be integrated with probabilistic inference to improve real-time machine learning prediction with accumulated prior predictions.

## CONCLUSIONS AND FUTURE WORK

Teaching neural networks to detect building objects from photos requires training; i.e., showing them many examples (up to thousands or even millions). Training datasets are often acquired by manually labeling data, which is labor-intensive and time-consuming. Ideas from the architectural field can be introduced to improve this process, specifically BIM and 3D rendering. BIM can generate synthetic data—as building object data—that can be used in conjunction with photorealistic renderings as training data in a deep learning neural network for understanding architectural elements in photos (or video images) about the built environment.

We presented valid methods for training-data synthesis using BIM and photorealistic rendering. Based on our results, we identify the following limitations and avenues for improvement: Generalization should be improved in the future by using more samples and training with more data. We assume that with enough data, we would be able to generalize trained networks on building photos for which we do not have corresponding training data (BIM models and renderings). Also, future work can explore using hyperreal parametric-BIM to automatically create large and fully synthetic datasets of photorealistic building models. As for the described semantic mesh model that was generated using photogrammetry-based texture creation, geometry processing methods would need to be developed to construct a valid BIM model in an appropriate file format.

The intended applications and significance of the methods described in this paper can be summarized in the following: Synthetic data generation methods provide virtually unlimited training data for deep learning. BIM and photorealistic rendering can generate high-quality training data for the applications of deep learning in architecture and will help overcome the bottleneck of labor-intensive data acquisition, especially in architecture where complex 3D and BIM data needed for training are already provided in the AEC industry. The same method can be applied to other industries where semantic models are available. As future work, the system presented in this paper can be augmented using hyperreal parametric-BIM and generative design methods to generate synthetic big data for training deep learning models to improve the generalization of the presented methods.

## IMAGE CREDITS

**Mohammad Alawadhi** is a doctoral student currently researching the intersection between artificial intelligence and architecture. He has an MSc in Architecture from Texas A&M. His interests include machine learning applications in architecture, computational design, and building energy modeling.

**Wei Yan**, Ph.D., Mattia Flabiano/Page Southerland Design Professor at Texas A&M University. He teaches computational methods in design and conducts research in BIM, simulation, optimization, AR, and AI, with projects funded by the NSF, NEH, DOE, Autodesk, etc.